\documentclass[letterpaper, 10 pt, conference]{ieeeconf}  % Comment this line out if you need a4paper
\IEEEoverridecommandlockouts                              % This command is only needed if
\usepackage{amsmath} % assumes amsmath package installed
\usepackage{amssymb}  % assumes amsmath package installed
\usepackage{cite} % sort and compress consecutive IEEE-style citations
\usepackage{booktabs}
\usepackage{array}
\usepackage{graphicx}
\usepackage{stfloats}
\usepackage{float}
\usepackage{multirow}
\usepackage{pifont}
\usepackage[table]{xcolor}
\makeatletter
\long\def\@makecaption#1#2{%
  \ifx\@captype\@IEEEtablestring
    \@IEEEtablecaptionsepspace%
    \parbox[t]{\hsize}{\centering\footnotesize\textbf{#1:} #2\par}%
    \vspace{0.35\baselineskip}%
  \else
    \@IEEEfigurecaptionsepspace%
    \setbox\@tempboxa\hbox{\footnotesize #1:\ #2}%
    \ifdim \wd\@tempboxa >\hsize
      \setbox\@tempboxa\hbox{\footnotesize #1:\ }%
      \parbox[t]{\hsize}{\footnotesize\noindent\unhbox\@tempboxa#2}%
    \else
      \ifcenterfigcaptions
        \hbox to\hsize{\footnotesize\hfil\box\@tempboxa\hfil}%
      \else
        \hbox to\hsize{\footnotesize\box\@tempboxa\hfil}%
      \fi
    \fi
  \fi
}
\makeatother

\definecolor{categoryrow}{RGB}{218,235,253}

\newcommand{\cmark}{\textcolor{green!60!black}{\ding{51}}}
\newcommand{\pmark}{\textcolor{black!50}{\ding{51}}}
\newcommand{\xmark}{\textcolor{red!75!black}{\ding{55}}}

\title{\LARGE \bf
DexBioLab:  Benchmarking Dexterous Manipulation with Dynamic Experimental Phenomena in Realistic Laboratories
}

\author{%
Xuanye Wu$^{1,2,*}$, Yue Shi$^{1,*,\dagger}$, Yongcan Yang$^{1,4}$,
Yiqun Wang$^{1,2}$, Zhaoqi Wan$^{1,2}$,\\
Dayu Xia$^{1}$, Yanming Shao$^{1}$, Jiezhang Cao$^{3}$, Xiangfei Li$^{2,\dagger}$,
Yao Mu$^{1,3,\dagger}$, Guangtao Zhai$^{1,3}$\\[0.5em]
{\normalsize $^{1}$Shanghai AI Laboratory\quad $^{2}$Huazhong University of Science and Technology}\\
{\normalsize $^{3}$Shanghai Jiao Tong University\quad $^{4}$University of Southern California}\\[0.3em]
{\footnotesize $^{*}$Equal contribution.\quad $^{\dagger}$Corresponding authors.}%
}

\begin{document}

\IEEEaftertitletext{%
  \vspace{0.2\baselineskip}%
  \refstepcounter{figure}%
  \label{fig:overview}%
  \centering
  \includegraphics[width=\textwidth]{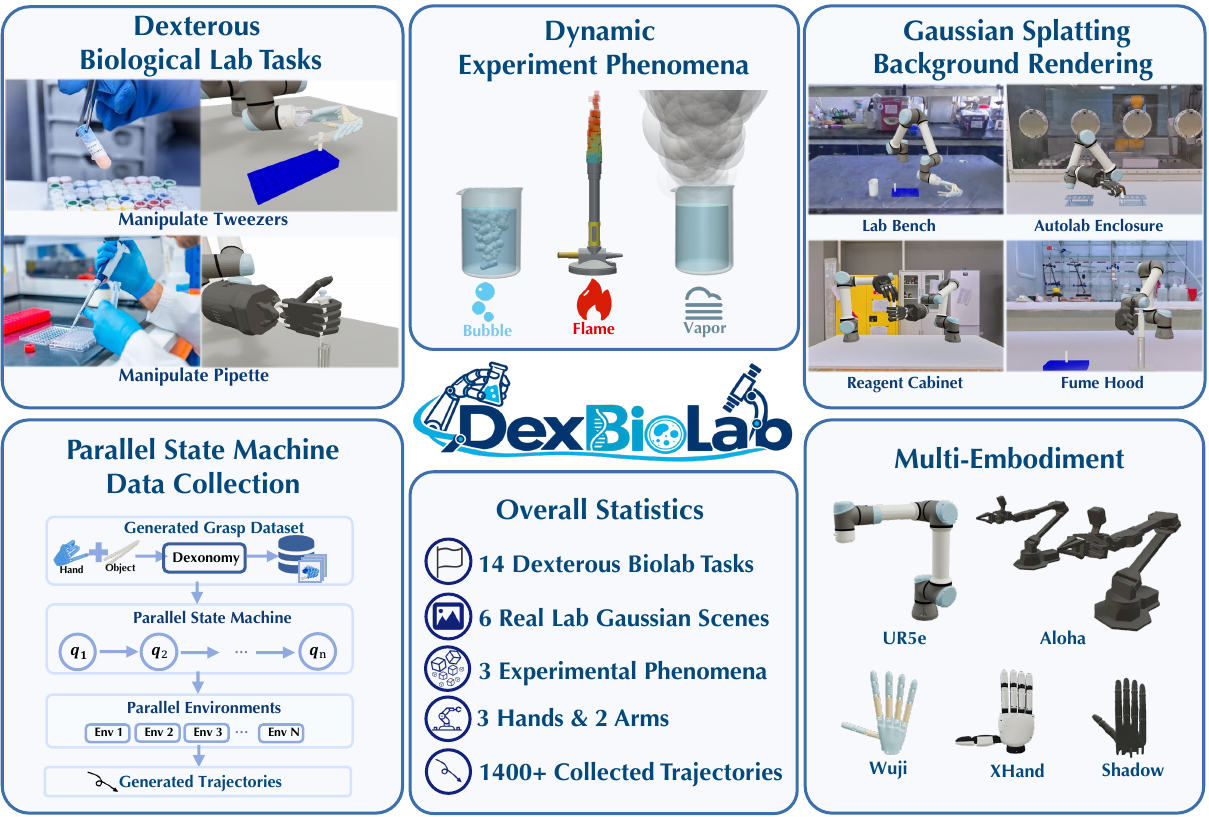}\\[0.3\baselineskip]
  \parbox{\textwidth}{\small
    Fig.~\thefigure: \textbf{Overview of DexBioLab.} DexBioLab is a benchmark for dexterous manipulation in biological laboratories, 
    comprising 14 tasks across four robot embodiments and incorporating three dynamic experimental phenomena, including flames, vapor, and bubbles. 
    Its hybrid rendering pipeline integrates interactive simulated foregrounds with 3D Gaussian Splatting (3DGS)~\cite{kerbl2023gaussians} backgrounds reconstructed from six real laboratory scenes,
    while parallel simulation supports efficient data generation and evaluation.}\\[0.6\baselineskip]
}

\maketitle
\thispagestyle{empty}
\pagestyle{empty}

%%%%%%%%%%%%%%%%%%%%%%%%%%%%%%%%%%%%%%%%%%%%%%%%%%%%%%%%%%%%%%%%%%%%%%%%%%%%%%%%

\begin{abstract}
Automating biological laboratories requires robots to perform dexterous tool interactions
under realistic visual conditions while handling dynamic experimental phenomena,
presenting significant challenges for robotic manipulation. To enable systematic
evaluation under these conditions, we introduce DexBioLab, a simulation benchmark and
framework for dexterous manipulation in realistic biological laboratory environments.
DexBioLab comprises 14 tasks across 4 robot embodiments, spanning functional tool use,
precision manipulation, language-conditioned visual grounding, and instrument operation.
We introduce controllable dynamic experimental phenomena, including flames, vapor, and
bubbles, to capture the evolving visual conditions of laboratory procedures. To further
enhance environmental realism, we develop a hybrid rendering pipeline that integrates
interactive simulated foreground objects with Gaussian-splatting backgrounds reconstructed
from 6 real laboratory scenes. For efficient large-scale data generation and evaluation,
we further build a parallel simulation framework on top of mjlab, proposing Parallel
Finite-State Machine (PFSM) and integrating batched rendering with decoupled trajectory replay
for demonstration generation and model-independent closed-loop policy evaluation. Our
evaluation reveals substantial variation in policy performance across tasks, exposing
distinct strengths and weaknesses of current manipulation policies. Real-world experiments
further demonstrate a reduced sim-to-real gap, supporting the validity of DexBioLab realistic manipulation evaluation. Supplementary materials are available at https://cumulusalpha.github.io/DexBioLabWebPage/.
\end{abstract}

\vspace{0.1cm}
\begin{table*}[t]
\centering
\caption{Comparison of DexBioLab with existing benchmarks. \cmark{} denotes full support, \pmark{} denotes partial support, and
\xmark{} denotes no support.}
\vspace{0.1cm}
\label{tab:benchmark_comparison}
\resizebox{\textwidth}{!}{
\begin{tabular}{lcccccc}
\toprule
\textbf{Benchmark}
& \textbf{Biolab}
& \textbf{Dexterous Manipulation}
& \textbf{Realistic Rendering}
& \textbf{Parallelization}
& \textbf{Multi-Embodiment}
& \textbf{Dynamic Phenomena}\\
\midrule
Bi-DexHands~\cite{chen2024bidexhands} & \xmark & \cmark & \xmark & \cmark & \xmark & \xmark\\
DexVerse~\cite{yao2026dexverse}       & \xmark & \cmark & \pmark & \pmark & \cmark & \xmark\\
DexJoCo~\cite{wang2026dexjoco}        & \xmark & \cmark & \xmark & \xmark & \xmark & \xmark\\

\midrule

Chemistry3D~\cite{li2025chemistry3d} & \cmark & \xmark & \pmark & \cmark & \xmark &\pmark\\
LabUtopia~\cite{li2025labutopia}     & \cmark & \xmark & \pmark & \xmark & \xmark & \xmark\\
Labimus~\cite{wu2026labimus} & \cmark & \cmark & \pmark & \xmark & \xmark & \xmark\\
AutoBio~\cite{lan2026autobio}        & \cmark & \pmark & \cmark & \xmark & \cmark & \xmark\\

\midrule

\textbf{DexBioLab (Ours)}
& \cmark
& \cmark
& \cmark
& \cmark
& \cmark 
& \cmark\\

\bottomrule
\end{tabular}
}
\end{table*}

%%%%%%%%%%%%%%%%%%%%%%%%%%%%%%%%%%%%%%%%%%%%%%%%%%%%%%%%%%%%%%%%%%%%%%%%%%%%%%%%
\section{Introduction}
Automated laboratories, often referred to as AutoLabs, are transforming scientific experimentation by shifting routine
procedures from manual operation toward autonomous execution~\cite{abolhasani2023selfdriving,king2009automation}. Conventional laboratory automation,
however, typically relies on task-specific instruments and tightly coupled workflows. Although such
systems are efficient in structured settings, adapting them to new experimental protocols, tools,
or operating conditions often requires substantial hardware modification and system reintegration.
General-purpose robotic manipulators equipped with dexterous hands or grippers provide a more flexible alternative, allowing
laboratory procedures to be adapted through demonstrations and learned policies rather than
hardware-specific re-engineering. This flexibility is particularly important in biological
laboratories, where robots must perform fine-grained operations such as manipulating pipettes and
tweezers, inserting tubes, actuating instruments, and coordinating multiple objects over
long-horizon procedures.

Recent robot-learning methods, particularly vision-language-action (VLA) policies,
have demonstrated strong manipulation capabilities in household 
and industrial environments~\cite{zhao2023learning,chi2023diffusion,black2025pi05}. 
To evaluate these learned policies, 
recent benchmarks have explored manipulation from several complementary perspectives. 
General-purpose manipulation benchmarks such as RoboTwin 2.0~\cite{chen2026robotwin2} 
and LIBERO~\cite{liu2023libero} provide diverse household and industrial tasks with 
controlled object interactions. Scientific robotics 
benchmarks~\cite{li2025chemistry3d,lan2026autobio,li2025labutopia} introduce 
laboratory-specific objects and procedures, while dexterous manipulation benchmarks such as 
DexJoCo~\cite{wang2026dexjoco}, DexVerse~\cite{yao2026dexverse}, and 
Bi-DexHands~\cite{chen2024bidexhands} focus on complex hand--object coordination. 
Recently, Labimus~\cite{wu2026labimus} extends dexterous manipulation evaluation 
to chemical laboratory scenarios. However, laboratory procedures often introduce temporally 
varying visual conditions, such as flames, vapor, and bubbles, which are underrepresented in 
existing manipulation benchmarks.

To address this gap, we introduce \textbf{DexBioLab}, 
a simulation benchmark for dexterous manipulation in realistic biological laboratory environments.
DexBioLab comprises 14 tasks across four robot embodiments, covering functional tool use, 
precision manipulation, language-conditioned visual grounding, and instrument operation. 
It incorporates controllable dynamic experimental phenomena, including flames, vapor, 
and bubbles, to capture visually evolving conditions during laboratory procedures. 
To improve visual realism while preserving physical interaction, DexBioLab proposes a hybrid 
rendering strategy that combines interactive simulated foreground objects with 
Gaussian-splatting backgrounds reconstructed from six real laboratory scenes. 
Leveraging mjlab as the simulation foundation, we design a parallel simulation 
framework centered around the Parallel Finite-State Machine (PFSM) policy, 
together with batched rendering and decoupled trajectory replay, 
enabling efficient demonstration generation and model-independent closed-loop evaluation.

Our contributions are summarized as follows:
\begin{itemize}
    \item \textbf{Multi-Embodiment Benchmark for Dexterous Laboratory Manipulation.}
    We introduce DexBioLab, comprising 14 laboratory manipulation tasks across four robot embodiments and covering functional tool use, precision manipulation, language-conditioned visual grounding, and instrument operation.
    \item \textbf{Dynamic Experimental Phenomena.}
    We incorporate controllable flames, vapor, and bubbles into laboratory procedures, introducing process-dependent visual changes for evaluating learned policies under
    temporally varying and perceptually complex observations.
    \item \textbf{Photorealistic Hybrid Rendering.}
    We combine interactive MuJoCo-rendered foreground objects with Gaussian-splatting backgrounds
    reconstructed from real laboratories, improving visual realism while preserving physical
    interaction and efficient batched rendering.
    \item \textbf{Parallel Data Generation and Evaluation.}
    Built on mjlab, DexBioLab supports parallel demonstration collection with the proposed PFSM policy, rendering, and closed-loop policy evaluation, enabling efficient large-scale data
    generation and benchmarking.

\end{itemize}

\section{Related Work}

\noindent\textbf{Visuomotor Policies.}
Visuomotor policies for robot manipulation have evolved from directly regressing actions
from visual observations toward increasingly structured generative policies. Early
Transformer-based approaches predict temporally coherent action chunks from visual observations,
improving the consistency of long-horizon execution~\cite{zhao2023learning}. Diffusion-based
imitation policies then model the action distribution through iterative denoising, providing a
flexible way to represent multimodal behaviors and contact-rich motions~\cite{chi2023diffusion,zhang2026r3dp}.
Building on these low-level visual policies, VLA models incorporate
large vision-language representations and language-conditioned action generation, enabling a single
policy to connect scene understanding, task instructions, and continuous control~\cite{black2025pi05}.
As VLA research has advanced, it has become increasingly clear that predicting how the world will
evolve under candidate actions can provide useful temporal and physical information for manipulation.
This insight has led to World Action Models (WAMs), which augment the VLA formulation by jointly
learning action generation and future visual-state prediction~\cite{bi2026motus,yuan2026fastwam}.
Despite these advances, existing visuomotor policies have been developed and evaluated predominantly
on general-purpose manipulation tasks. Their capabilities in scientific laboratory settings,
particularly for the fine-grained, contact-rich, and dexterous operations required in biological
experiments, remain insufficiently explored.
Our benchmark is designed to address the lack of evaluation for dexterous manipulation in
photorealistic biological laboratory settings, featuring a diverse set of tasks that capture
fine-grained operations.

\noindent\textbf{Manipulation Benchmarks.}
Existing manipulation benchmarks commonly evaluate general robotic capabilities in household or
industrial environments.
Benchmarks such as RoboTwin 2.0~\cite{chen2026robotwin2}, RLBench~\cite{james2020rlbench}, and
BEHAVIOR-1K~\cite{li2024behavior1k} cover a wide variety of tasks, and select parallel-jaw grippers as their robot embodiments. 
The low-dimensional control space of these grippers supports efficient trajectory collection and
large-scale evaluation, yet it can hardly fully represent the fine-grained coordination required for
human-like manipulation.
In contrast, dexterous manipulation benchmarks such as Bi-DexHands~\cite{chen2024bidexhands},
DexVerse~\cite{yao2026dexverse}, and DexJoCo~\cite{wang2026dexjoco} emphasize multi-fingered hand
control and complex, contact-rich interactions.
They provide valuable testbeds for studying coordinated finger motion, bimanual control, and
generalization across objects, tasks, or robot embodiments. 
These benchmarks provide valuable foundations for studying dexterous manipulation, but laboratory-specific procedures involving diverse tools, instruments, and dynamic experimental phenomena remain less explored. Meanwhile, recent simulation frameworks have investigated the integration of neural rendering with interactive physics simulators to improve visual realism while preserving physical interaction~\cite{jia2025discoverse}. However, such simulation advances have not yet been integrated into benchmarks for evaluating dexterous laboratory manipulation under realistic and evolving experimental conditions.
% These benchmarks primarily focus on general manipulation settings, with limited coverage of biological laboratory scenarios and the dynamic experimental phenomena that shape perception and decision-making in real experimental workflows. DexBioLab complements these general benchmarks by placing dexterous manipulation within diverse biological laboratory settings. Its tasks are structured around realistic experimental operations, allowing robotic models to be evaluated not only on low-level hand control but also on their ability to complete purposeful, multi-stage laboratory procedures.

\noindent\textbf{Scientific Robotics Benchmarks.}
Recent scientific robotics benchmarks have begun to investigate robotic manipulation in laboratory
environments. Chemistry3D~\cite{li2025chemistry3d} and LabUtopia~\cite{li2025labutopia} provide
scientific manipulation scenarios, but primarily focus on gripper-based interactions.
AutoBio~\cite{lan2026autobio} extends this line of work to dexterous manipulation with a limited
set of laboratory tasks, while simplifying hand control to binary open--close actions.
In a related direction, Labimus~\cite{wu2026labimus} studies dexterous manipulation by humanoid
robots in chemical laboratories.
Against this backdrop, DexBioLab broadens the scope of scientific robotics benchmarking with a
larger and more diverse task suite, multiple robot embodiments, and richer dynamic visual effects,
enabling a more comprehensive evaluation of robot-learning models.

\section{DexBioLab Simulation}
To support realistic parallel evaluation in biological laboratory environments, we develop
\textbf{DexBioLab Simulation}, a photorealistic and parallelizable simulation framework built on
mjlab~\cite{zakka2026mjlab}. Our approach extends the existing mjlab infrastructure with enhanced rendering
capabilities, dynamic experimental phenomena, and system-level parallelization. This
design preserves the robustness and physical accuracy of the underlying physics engine while
addressing its limitations in visual realism within biolab scenarios.

The framework comprises three key components: (A) a particle visual-effects system, (B) hybrid rendering, and (C) parallel simulation.

\subsection{Particle System for Dynamic Experimental Phenomena}
\looseness=-1
Biological laboratory environments often contain dynamic visual phenomena, such as visible vapor
formation and time-varying flame intensity, which are perceptually important but expensive to model with full physics. To preserve scalability, we propose a \textbf{particle-based visual effect system} that
favors perceptual fidelity while remaining lightweight and parallelizable. Each visual element is a
particle primitive with state
$\mathbf{s}_i(t)=(\mathbf{x}_i(t),\mathbf{c}_i(t),\alpha_i(t),\sigma_i(t))$. Here,
$\mathbf{x}_i(t)$ denotes position, while $\mathbf{c}_i(t)$, $\alpha_i(t)$, and $\sigma_i(t)$ specify
the particle color, opacity, and visual size, respectively. All particle states are decoupled from
the scene physics and updated purely for rendering.

\begin{figure*}[t]
    \centering
    \includegraphics[width=\textwidth]{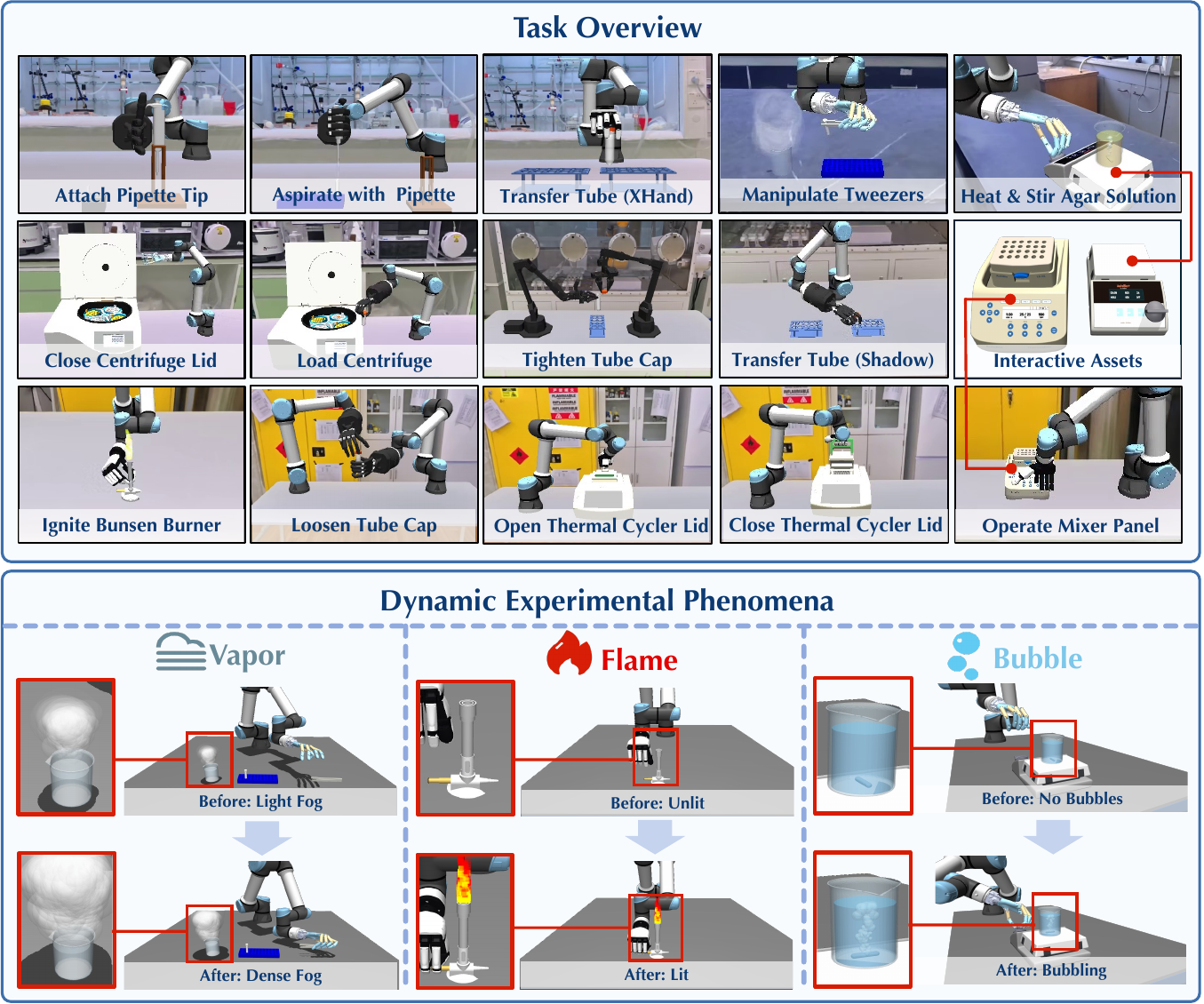}
    \caption{\textbf{Tasks Overview and Dynamic Experimental Phenomena.} The top figure illustrates all 14 tasks in DexBioLab along with the interactive assets, while the bottom figure presents the 3 types of dynamic experimental phenomena in DexBioLab, highlighting their visual evolution and temporal variations.}
    \label{fig:task-overview}
\end{figure*}

At each time step of duration $\Delta t$, particle positions are updated using a lightweight rule
that combines deterministic drift with a stochastic perturbation:
\begin{equation}
\mathbf{x}_i(t+\Delta t) = \mathbf{x}_i(t)
    + \left[\mathbf{d}_i(\mathbf{x}_i,t) + \boldsymbol{\eta}_i(t)\right]\Delta t,
\end{equation}
where $\mathbf{d}_i(\mathbf{x}_i,t)$ is a deterministic drift field that encodes effects such
as buoyancy and upward motion, and $\boldsymbol{\eta}_i(t)$ is a small random displacement-rate
perturbation that produces flicker and irregular motion. The scale $\sigma_i(t)$ may be varied over
the particle lifetime to model expansion or growth. Particle opacity decays with age (e.g.,
$\alpha_i(t)=\alpha_{0, i}\exp(-(t-t_{0, i})/\tau_i)$, \text{where} $\tau_i$ is a predefined time constant), and particles are spawned from localized sources
with controlled emission rates and parameter distributions, including initial scales. The design is
inspired by real-time game-engine particle systems~\cite{reeves1983particle}, producing perceptually
convincing vapor, flame, and bubbles at low cost while remaining decoupled from the
physics simulation. The effects are illustrated in the lower panel of
Fig.~\ref{fig:task-overview}.

\noindent\textbf{Flame.}
The flame is approximated by upward-moving cube particles with temporally varying color and opacity. The
particles follow stochastic upward drift, exhibit gradual alpha decay, and use color gradients to
convey high-temperature cores and cooler edges. The flame color is also
modulated by the rotational positions of the Bunsen burner's two rings, causing its color to vary with
burner adjustment.

\noindent\textbf{Vapor.}
For phenomena such as the visible vapor produced by liquid nitrogen in a container, we model
vapor as expanding particle clouds with randomized drift and dispersion. 
The spherical particles increase their spread over time, 
modulate their opacity based on particle age and local density, 
and respond to coarse airflow to indicate directional drift.

\noindent\textbf{Bubbles.}
To represent bubble formation and stirring-induced motion, such as during heating or magnetic
stirring, we simulate rising particles with vortex-like trajectories and varying rise speeds. The
particles combine upward buoyant motion with rotational offsets to approximate vortices and bubble
paths.

% \subsection{Photorealistic Gaussian Rendering}
\subsection{Hybrid Rendering for Realistic Laboratory Environments}
To achieve realistic laboratory environments while preserving physical interaction, we develop a hybrid rendering approach that integrates Gaussian-splatting backgrounds with simulator-rendered interactive foreground objects. Unlike replacing the simulator renderer entirely, our approach maintains the physical controllability of simulated objects while introducing realistic visual contexts.
% Inspired by the hybrid 3D Gaussian Splatting and simulator design of DISCOVERSE~\cite{jia2025discoverse},
% we combine Gaussian-rendered backgrounds with simulator-rendered foregrounds rather than replacing the original renderer.
Some of the laboratory assets are sourced from AutoBio~\cite{lan2026autobio} and LabUtopia~\cite{li2025labutopia}.
The 3DGS assets for six laboratory scenes are captured with a lightweight spatial scanner (Lixel K2)\footnote{\textnormal{https://www.xgrids.com/intl/lixelk2}}. The final image is formed as
\begin{equation}
    \begin{aligned}
    I_{\mathrm{final}} &= M \odot I_{\mathrm{origin}} \\
    &\quad + (1-M) \odot
    \mathcal{R}\!\left(\{G_i\}_{i=1}^{N},T_{\mathrm{cam}}\right),\\
    &\qquad M\in\{0,1\}^{H\times W}.
    \end{aligned}
\end{equation}
Here, $M$ is mjlab's object-level foreground mask, $I_{\mathrm{origin}}$ is the original simulator
image, $\{G_i\}_{i=1}^{N}$ are the preconstructed scene Gaussians, $T_{\mathrm{cam}}$ is the camera
pose, $\mathcal{R}$ is the Gaussian splatting renderer, and $\odot$ denotes element-wise
multiplication. This hybrid design
retains interactive simulator objects while providing efficient, photorealistic laboratory backgrounds.

\begin{figure*}[t]
    \centering
    \includegraphics[width=\textwidth]{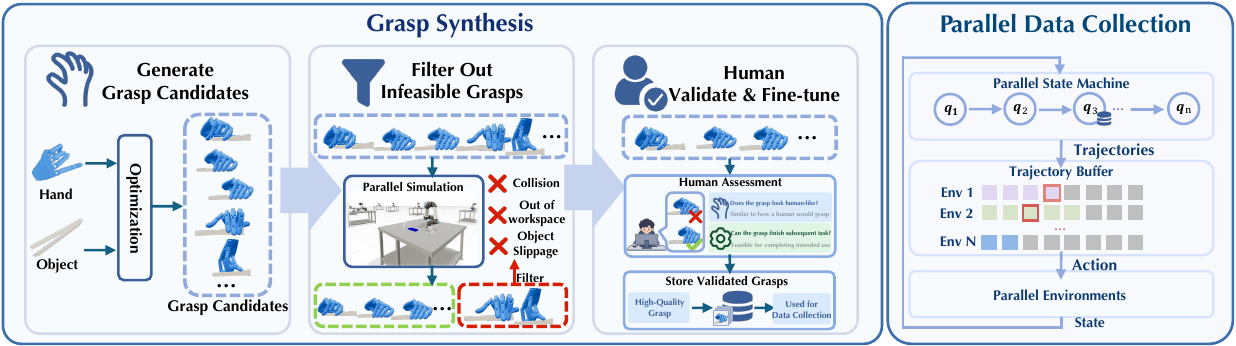}
   \caption{
    % \textbf{Grasp Synthesis and Parallel Finite State Machine (PFSM) for Data Collection.}
\textbf{Grasp Synthesis.} First, large-scale grasp candidates are generated via optimization-based methods (e.g., Dexonomy and Lightning Grasp). 
Then, parallel simulation rollout is performed, and candidates are filtered based on feasibility constraints such as workspace limits and collisions. 
Finally, manual verification is conducted to ensure human-like plausibility. The validated grasps are stored in a grasp dataset.
\textbf{PFSM Policy:} A PFSM operates over all environments in a batched manner, where each environment corresponds to an individual state instance. 
Transitions are triggered based on observed states, and planned trajectories are stored in a buffer that outputs batched actions when required.}
    \label{fig:dataset-generation-pipeline}
\end{figure*}

\subsection{Parallel Simulation}\nopagebreak
A primary goal of DexBioLab Simulation is to enable parallel data generation and evaluation for
modern robot-learning methods. To achieve this, the system is designed with parallelism as a core
principle, allowing rendering, visual effects, and environment simulation to run efficiently across many instances.

The Gaussian rendering pipeline supports batched rendering, allowing multiple environments to be
processed simultaneously on the GPU. This vectorized design significantly improves throughput
compared with sequential rendering. The particle-based visual effects are also naturally
parallelizable because individual particles can be updated independently without global
synchronization.

We further support texture variability by extending AutoBio's interactive-texture framework.
Because Warp does not support arbitrary texture changes at runtime, we predefine a set of textures
in the MJCF scene description and index them during simulation for efficient swapping. This design
extends AutoBio's support for interactive textures while remaining compatible with parallel
execution and interactive instruments in the scene.
Together, these components form a unified parallel simulation pipeline capable of scaling to a large
number of environments. In a representative benchmark using 339 frames from two cameras at a
resolution of $640 \times 480$, plain MuJoCo Warp rendering required 13.95~s, Gaussian-augmented rendering
required 15.57~s, and Blender rendering used by AutoBio required 1092.02~s. Thus, DexBioLab rendering is
approximately $70.14\times$ as fast as AutoBio while incurring approximately $11.61\%$ overhead
compared with plain MuJoCo Warp.
This overhead represents an acceptable tradeoff given the visual benefits and the substantial speed
advantage over AutoBio workflows.

\section{DexBioLab Benchmark}
\subsection{Design Principles for Tasks}
\label{sec:task-design-principles}
The top panel of Fig.~\ref{fig:task-overview} summarizes the task suite and the simulation features that support
interactive assets. The DexBioLab task design follows four principles intended to represent the
manipulation capabilities required in biological laboratories rather than generic object
rearrangement. (1) \textbf{Functional Tool
Use.} Tasks require a policy to exploit the intended function of a laboratory tool rather than
merely grasping and relocating it. (2) \textbf{Precision Manipulation.}
Successful task execution requires precise manipulation and coordinated finger motion during
contact-rich interactions. (3) \textbf{Language-Conditioned Visual Grounding.} A policy must associate language instructions 
with spatial entities in the current observation and act on the visually identified target. 
For example, a robot may be required to insert a tube into a rack at a location specified in the instruction, such as “row 2, column 3”. 
This requires grounding language descriptions into precise spatial 
targets and jointly reasoning over language and scene geometry. 
(4) \textbf{Instrument Operation.} Tasks involve interaction with laboratory instruments through articulated controls and physical manipulation, requiring the policy to operate devices as part of the task execution. These scenarios are commonly encountered in biological laboratories.
Some of the tasks are adapted from AutoBio~\cite{lan2026autobio} and redesigned for dexterous hands.

\subsection{Dataset Generation}

\noindent\textbf{Grasp Synthesis.}
Before trajectory synthesis, we generate task-specific grasp candidates from the
object geometry and the kinematic structure of the hand.
Learning-based methods can directly generate grasp poses
~\cite{wei2024dro,fei2025trograspefficientgraph}, but require training on grasp datasets.
Hence, we use Dexonomy~\cite{chen2025dexonomy} and Lightning Grasp~\cite{yin2025lightninggrasp} 
to generate a wider variety of grasp types.
The resulting hand poses are represented in
the object frame and converted into embodiment-specific pre-grasp and grasp targets for
subsequent trajectory planning.

We instantiate the generated grasp candidates in the parallel physics simulator and
evaluate them by replaying the transition from a pre-grasp configuration to the
final grasp. We filter out candidates that are not reachable by the manipulator,
violate collision constraints with the environment, or fail to maintain stable
object contact under subsequent motion. In particular, we reject grasps that lie
outside the robot workspace, induce hand--environment collisions during execution,
or lead to object slippage when the arm moves. This stage ensures that the retained
candidates are both kinematically feasible and dynamically stable under the contact
dynamics used for demonstration collection.

Finally, we incorporate a human validation stage to refine the remaining
candidates further. While some grasps are physically feasible, they may exhibit unnatural
or implausible hand configurations, such as self-intersecting fingers or awkward
poses that deviate from typical human grasping strategies and hinder subsequent
task execution. We therefore perform
manual inspection to filter out such cases, retaining grasps that are both
physically valid and human-like. This step improves the quality and realism of the
resulting demonstrations by aligning them with human manipulation priors. Fig.~\ref{fig:dataset-generation-pipeline}
(left) illustrates the overall grasp synthesis pipeline.

\noindent\textbf{PFSM Policy for Data Collection.}
We generate demonstrations using task-specific scripted policies implemented as
parallel finite-state machine. Each task is decomposed into a sequence of semantic stages 
(e.g., approaching and grasping), where each stage is represented by a set of end-effector waypoints and corresponding hand or gripper joint configurations. 
Across $N$ parallel environments, each environment independently maintains its current stage index $q_i^t$ and waypoint index $k_i^t$. Given observation $\mathbf{o}_i^t$,
the corresponding controller produces
\begin{equation}
    \mathbf{a}_i^t
    =
    \pi_{q_i^t}\!\left(\mathbf{o}_i^t, k_i^t\right),
    \qquad
    \mathbf{A}^t
    =
    [\mathbf{a}_1^t,\ldots,\mathbf{a}_N^t].
\end{equation}
After each simulation step, the local state is updated as
\begin{equation}
(q_i^{t+1},k_i^{t+1})
=
\begin{cases}
F_{q_i^t}(q_i^t, k_i^t, \mathbf{o}_i^{t+1}),
& c_{q_i^t}(\mathbf{o}_i^{t+1}, k_i^t)=1,\\
(q_i^t, k_i^t),
& \text{otherwise},
\end{cases}
\end{equation}
where $c_q$ determines whether the current waypoint or stage is complete, and
$F_q$ advances the corresponding state. Since this update is performed
independently for each environment, different task stages can be executed
asynchronously within the same batched simulation. The PFSM policy data collection pipeline is illustrated on the right of Fig.~\ref{fig:dataset-generation-pipeline}.

To further improve throughput, we decouple physics rollout and image rendering.
During rollout, camera rendering is disabled, and each environment stores its
state, action, timestamp, and language goal in an independent
buffer $D_i$. Only successful trajectories are retained and replayed
in batches to render multi-view RGB observations and then packaged into datasets. This procedure
avoids rendering during trajectory search while preserving frame-level
alignment between control and visual observations. 

\section{Experiments}

\subsection{Simulation Benchmark Evaluation}
We evaluate four representative policies---Diffusion Policy, $\pi_{0.5}$, GR00T-N1.7, and
Fast-WAM---on 12 representative tasks selected from the full 14-task DexBioLab suite. The
selection is stratified across the four capability categories and includes all four robot
embodiments. All models are trained on the same task-specific dataset and evaluated in the
DexBioLab simulator, using identical domain-randomization settings, and task-specific termination criteria. For each selected task, we report the mean
success rate and SEM across 3 seeds (300 rollouts in total, with 100 rollouts per seed).
All tasks use binary rewards except Operate Mixer Panel,
which uses a progress reward.
The resulting task-level comparison is reported in Table~\ref{tab:task_success_rates} and organized
according to the task categories introduced in Sec.~\ref{sec:task-design-principles}.
\begin{table*}[!t]
\centering
\caption{Task-wise success rates (\%) on the 12 representative DexBioLab tasks: mean $\pm$ SEM over 3 seeds, 100 rollouts per seed, 300 rollouts total.}
\label{tab:task_success_rates}
\small
\renewcommand{\arraystretch}{1.08}
\setlength{\tabcolsep}{5pt}
\begin{tabular*}{\textwidth}{@{\extracolsep{\fill}}lcccc@{}}
\toprule
\textbf{Task}
& \textbf{DP}
& \textbf{$\boldsymbol{\pi}_{\mathbf{0.5}}$}
& \textbf{GR00T-N1.7}
& \textbf{Fast-WAM} \\
\midrule
\rowcolor{categoryrow}
\multicolumn{5}{c}{\textit{Functional Tool Use}} \\
Aspirate with Pipette (Shadow)                 & $0.3 \pm 0.3$ & $0.0 \pm 0.0$ & $1.0 \pm 0.0$ & $\mathbf{4.7 \pm 1.5}$ \\
Manipulate Tweezers (Wuji)                  & $0.7 \pm 0.3$ & $\mathbf{6.7 \pm 2.0}$ & $4.0 \pm 1.7$ & $1.0 \pm 0.6$ \\
\midrule
\rowcolor{categoryrow}
\multicolumn{5}{c}{\textit{Precision Manipulation}} \\
Tighten Tube Cap (Aloha)             & $0.0 \pm 0.0$ & $\mathbf{0.7 \pm 0.7}$ & $0.0 \pm 0.0$ & $0.0 \pm 0.0$ \\
Attach Pipette Tip (Shadow)          & $5.0 \pm 1.2$ & $12.0 \pm 0.0$ & $2.7 \pm 0.7$ & $\mathbf{20.7 \pm 1.8}$ \\
\midrule
\rowcolor{categoryrow}
\multicolumn{5}{c}{\textit{Language-Conditioned Visual Grounding}} \\
Load Centrifuge (Shadow) & $0.0 \pm 0.0$ & $\mathbf{1.0 \pm 0.6}$ & $0.3 \pm 0.3$ & $0.7 \pm 0.3$ \\
Transfer Tube (Shadow)       & $2.7 \pm 0.9$ & $1.3 \pm 0.3$ & $\mathbf{12.0 \pm 1.5}$ & $0.0 \pm 0.0$ \\
Operate Mixer Panel (XHand)              & $0.0 \pm 0.0$ & $5.5 \pm 2.0$ & $\mathbf{5.7 \pm 1.4}$ & $0.0 \pm 0.0$ \\
\midrule
\rowcolor{categoryrow}
\multicolumn{5}{c}{\textit{Instrument Operation}} \\
Ignite Bunsen Burner (XHand)        & $60.7 \pm 5.4$ & $\mathbf{77.7 \pm 4.8}$ & $40.0 \pm 7.4$ & $23.3 \pm 3.7$ \\
Close Centrifuge Lid (Wuji)          & $30.0 \pm 0.6$ & $81.7 \pm 1.2$ & $\mathbf{100.0 \pm 0.0}$ & $43.3 \pm 29.7$ \\
Heat \& Stir Agar Solution (Wuji)     & $1.7 \pm 0.7$ & $\mathbf{7.3 \pm 1.8}$ & $\mathbf{7.3 \pm 2.9}$ & $2.3 \pm 0.9$ \\
Close Thermal Cycler Lid (XHand)         & $87.0 \pm 1.0$ & $5.3 \pm 0.9$ & $49.7 \pm 0.7$ & $\mathbf{99.7 \pm 0.3}$ \\
Open Thermal Cycler Lid (XHand)          & $14.3 \pm 2.2$ & $\mathbf{56.7 \pm 1.2}$ & $18.0 \pm 4.0$ & $5.7 \pm 3.7$ \\
\bottomrule
\vspace{0.01cm}
\end{tabular*}
% \vspace{0.7em}
% Keep the two real-world panels inside the same double-column float as the table.
\begin{minipage}[t]{0.48\textwidth}
    \centering
    \includegraphics[width=\linewidth]{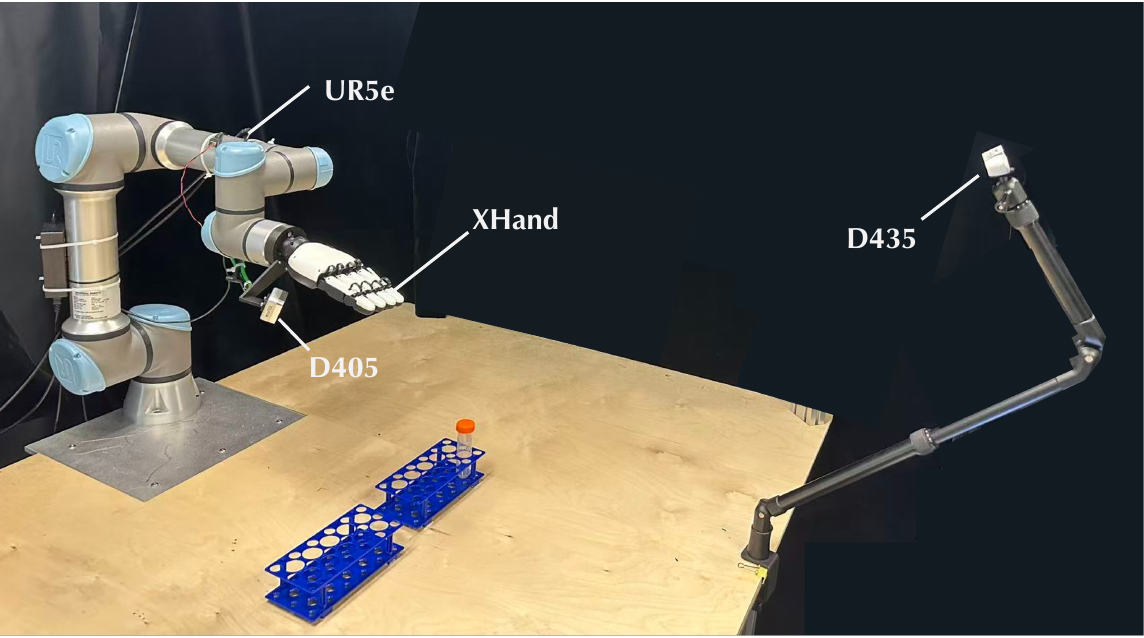}
    \refstepcounter{figure}\label{fig:realworld_setup}
    \parbox[t]{\linewidth}{\footnotesize\textbf{Fig.~\thefigure: Real-world setup.}
    % A UR5e arm with an XHand is used, along with a RealSense camera system consisting of a wrist-mounted D405 and a third-view D435 providing observation of the workspace.
    A UR5e manipulator equipped with an XHand is used together with a RealSense camera system, including a wrist-mounted D405 and a third-person D435, to provide complementary observations of the workspace during policy evaluation.}
\end{minipage}\hfill
\begin{minipage}[t]{0.48\textwidth}
    \centering
    \includegraphics[width=\linewidth]{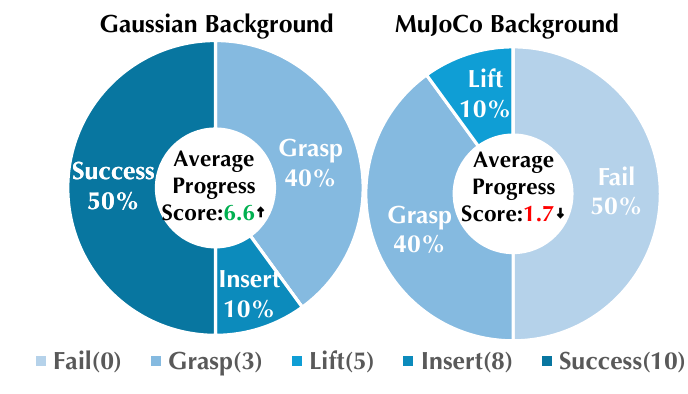}
    \refstepcounter{figure}\label{fig:realworld_score_distribution}
    \parbox[t]{\linewidth}{\footnotesize\textbf{Fig.~\thefigure: Progressive Score Distribution.} The pie charts compare the distribution of task progress across 20 real-world trials under Gaussian and MuJoCo backgrounds, showing that Gaussian backgrounds lead to better performance and average scores.}
\end{minipage}
\vspace{-0.8em}
\end{table*}

% \iffalse
% \vspace{-1.8em}
% \subsection{Evaluation Infrastructure}
% DexBioLab provides a unified evaluation infrastructure built around domain randomization,
% task-specific success checking, and model-independent policy communication. At each episode reset,
% DexBioLab applies domain randomization to the robot's initial joint configuration and the object poses.
% Each task is equipped with a task-specific success-checking function that determines completion from the simulator state.
% The DexBioLab and policy runtime are decoupled via a ZeroMQ\cite {hintjens2013zeromq} observation–action interface, allowing policies to run in independent Python environments with different versions and dependencies. The
% simulator serializes the current visual observations, proprioceptive state, and language prompt to
% an external inference process and receives action chunks in return. Consequently, a model can be
% evaluated from a separate process, machine, or software environment without being integrated into
% the simulator's dependency stack.
% \fi

\subsection{Real-World Validation}
To examine whether DexBioLab provides observations that are aligned with real-world data and meaningful for evaluating manipulation policies, 
we conduct a real-world validation using a UR5e robot arm equipped with an XHand and a camera system, 
consisting of an Intel RealSense D405 mounted on the wrist and an Intel RealSense D435 providing a third-person view, as shown in Fig.~\ref{fig:realworld_setup}.

We use a few-shot data mixture to examine whether the benchmark's simulated data capture
real-world-relevant visual cues. Each training set contains 10 real-robot demonstrations and 90 simulated
demonstrations. The two conditions differ only in the simulated visual background: the
\emph{Gaussian-background} condition renders the simulated demonstrations with Gaussian-splatting
backgrounds, whereas the \emph{MuJoCo-Background} condition uses the standard mjlab background.
We evaluate both conditions on the same real-robot task using a progressive score. The scoring
rubric is: 3 points for grasping the tube, 5 points for successfully picking it up, 8 points for
initiating insertion, and 10 points for fully inserting it into the rack. The tube is placed at
each of ten positions in the initial rack, and each position is tested twice, resulting
in 20 trials per condition. As summarized in Fig.~\ref{fig:realworld_score_distribution}, the
\emph{Gaussian-background} condition achieves a normalized mean score of 66\%, compared with 17\%
without Gaussian backgrounds. This gap indicates that realistic background appearance can be
particularly important in the few-shot regime, where the limited number of real demonstrations
cannot by themselves characterize the full visual variation of the task. These results support
the fidelity and practical relevance of the benchmark's Gaussian-background data for real-world
policy evaluation.

\section{Discussion}

\noindent\textbf{Language-conditioned visual grounding remains a major bottleneck.}
Across language-conditioned visual grounding tasks, success rates remain
generally low and highly variable across policies, and none of the tasks is
solved reliably. Some models even achieve 0\% success on certain tasks. These tasks require policies to ground language-specified spatial or symbolic targets in the visual scene, such as selecting a tube slot by row and column or identifying a target value on an instrument panel. The
results suggest that current policies still struggle to combine language-based
target identification with spatial reasoning and precise motion control.
Notably, Fast-WAM does not consistently outperform the VLA baselines on these
grounding tasks, suggesting that the main limitation may lie less in predicting
future visual states and more in the language-conditioned front end. Improving
language grounding and instruction-conditioned scene representations may
therefore be necessary before the additional world-modeling capacity of WAMs
can translate into better performance.

\noindent\textbf{Dexterous-hand pretraining alone does not ensure robust task performance.}
Performance varies substantially across tasks requiring dexterous-hand control.
Dexterous-hand pretraining appears beneficial for relatively simple or
tolerance-rich interactions. For example, GR00T-N1.7 achieves 100\% success on
Close Centrifuge Lid. However, this benefit does not consistently extend to
contact-rich tasks that require precise alignment, fingertip coordination, and
continuous finger adjustment. GR00T-N1.7 remains weak on pipette, tweezer, and
tube-cap manipulation, despite involving dexterous-hand control. These results
suggest that pretraining can provide a useful initialization for basic
manipulation skills, but task-specific demonstrations and contact-aware control
are still necessary for reliable execution of laboratory operations.

% \noindent\textbf{Contact-rich operations constitute the main gap for laboratory automation.}
\noindent\textbf{Precise contact-rich manipulation remains a prominent failure mode.}
Across the evaluated tasks, visuomotor policies achieve lower success rates on contact-rich tasks, such as manipulating tweezers, tightening tube caps, attaching pipette tips, loading centrifuges, and transferring tubes, but perform better on tasks involving nonprehensile interactions, 
such as closing centrifuge lids and opening thermal cycler lids. 
% This gap suggests that improving contact-rich control remains essential before VLA policies can be reliably deployed in laboratories, where such operations are ubiquitous.
These results highlight the importance of improving contact-aware control and
fine-grained visuomotor coordination for reliable laboratory manipulation.

\section{Limitations and Future Work}

\noindent\textbf{Task and asset scalability.}
Although DexBioLab encompasses a wide range of dexterous laboratory manipulation tasks, it does not yet provide a framework for evaluating the full scope of biological laboratory automation. Future work could employ generative approaches to synthesize assets and tasks, thereby reducing manual effort and broadening the scope and content of evaluations for biological experiments.

\noindent\textbf{Rendering and dynamic-scene fidelity.}
While DexBioLab renders biolab scenes using Gaussian-background compositing together with a particle-based system, it may not fully reproduce high-fidelity liquid behavior or the full range of scene interactions. Future work could explore 4D Gaussian representations or dynamic neural fields to better capture liquid dynamics and scene interactions.

\section{Conclusion}
This paper introduced DexBioLab, a benchmark for dexterous manipulation evaluation in realistic laboratory environments with dynamic experimental phenomena.
DexBioLab comprises 14 tasks across four robot embodiments and four capability
categories, and incorporates controllable flames, vapor, and bubbles to represent
process-dependent visual changes. To provide realistic
visual context, we develop a hybrid rendering
framework which combines 3D Gaussian
Splatting backgrounds reconstructed from real laboratories. We further build a
parallel simulation framework, introducing Parallel Finite-State Machine (PFSM) for efficient
demonstration generation and policy evaluation.
Extensive evaluations reveal substantial variation in policy performance across tasks
and expose persistent limitations in contact-rich interaction, language--spatial
grounding, and fine-grained dexterous manipulation. Real-world experiments further
show that Gaussian-splatting backgrounds reduce the sim-to-real gap and improve few-shot transfer. We hope DexBioLab can facilitate future research on robust dexterous manipulation and
perception in laboratory environments.
% This paper presented DexBioLab, a benchmark for evaluating the dexterous
% manipulation capabilities of current visuomotor policies in realistic
% laboratory environments with dynamic experimental phenomena. DexBioLab
% models flames, vapor, and bubbles as dynamic experimental phenomena. Its
% hybrid renderer combines two complementary types of scene assets, interactive
% simulator-rendered foreground objects and 3D Gaussian Splatting backgrounds 
% to create realistic, visually dynamic scenes while retaining efficient batched
% simulation. Built on mjlab, the framework further integrates the proposed PFSM policy with batched rollout
% to accelerate large-scale demonstration collection and policy evaluation. The proposed
% benchmark comprises 14 tasks across four complementary capability categories and
% four robot embodiments. Our simulation experiments show that current policies remain limited in contact-rich interactions,
% language--spatial grounding, and fine-grained dexterous manipulation, highlighting
% these capabilities as important directions for future research toward robust
% laboratory automation. Real-world validation further examines whether DexBioLab
% observations are aligned with real-world laboratory data and meaningful for
% evaluating manipulation policies. In the few-shot sim-to-real setting,
% Gaussian-splatting backgrounds substantially improve transfer performance over
% standard MuJoCo backgrounds, supporting the visual fidelity and practical
% relevance of DexBioLab for real-world laboratory scenarios.

\section*{Acknowledgment}
This work was supported by New Generation Artificial Intelligence-National Science and Technology Major Project (2025ZD0124104) in collaboration with Shanghai Artificial Intelligence Laboratory.
We thank the Laboratory of Bioheat and Mass Transfer, School of Biomedical Engineering, Shanghai Jiao Tong University, for providing the biological laboratory scene used in the Manipulate Tweezers task.
We also extend our sincere thanks to Liang Zhiqian, Jiang Yuxuan, Shao Yanming, and Liang Liang for their technical support, which greatly aided the development and refinement of this work.

\bibliographystyle{ieeetr}
\bibliography{references}

\begin{thebibliography}{10}

\bibitem{kerbl2023gaussians}
B.~Kerbl, G.~Kopanas, T.~Leimk{\"{u}}hler, and G.~Drettakis, ``3d gaussian
  splatting for real-time radiance field rendering,'' {\em ACM Transactions on
  Graphics}, vol.~42, no.~4, 2023.

\bibitem{chen2024bidexhands}
Y.~Chen, Y.~Geng, F.~Zhong, J.~Ji, J.~Jiang, Z.~Lu, H.~Dong, and Y.~Yang,
  ``{Bi-DexHands}: Towards human-level bimanual dexterous manipulation,'' {\em
  IEEE Transactions on Pattern Analysis and Machine Intelligence}, vol.~46,
  no.~5, pp.~2804--2818, 2024.

\bibitem{yao2026dexverse}
Y.~Yao, Z.~Xu, T.~Zhang, Z.~Liu, S.~Li, Z.~Wei, F.~Chen, D.~Huang, K.~Wan,
  C.~Ma, S.~Zhao, S.~Gao, M.~Tomizuka, Y.~Ma, and M.~Ding, ``{DexVerse}: A
  modular benchmark for multi-task, multi-embodiment dexterous manipulation,''
  {\em arXiv preprint arXiv:2607.08751}, 2026.

\bibitem{wang2026dexjoco}
H.~Wang, W.~Zhao, X.~Wang, S.~Huang, H.~Lin, B.~Zheng, R.~Xu, G.~Wang, Y.~Mu,
  H.~Wang, L.~Fan, H.~Li, Z.~Zhang, and T.~Tan, ``{DexJoCo}: A benchmark and
  toolkit for task-oriented dexterous manipulation on {MuJoCo},'' {\em arXiv
  preprint arXiv:2605.16257}, 2026.

\bibitem{li2025chemistry3d}
S.~Li, Y.~Huang, C.~Guo, T.~Wu, J.~Zhang, L.~Zhang, and W.~Ding,
  ``{Chemistry3D}: Robotic interaction toolkit for chemistry experiments,'' in
  {\em 2025 IEEE International Conference on Robotics and Automation}, 2025.

\bibitem{li2025labutopia}
R.~Li, Z.~Hu, W.~Qu, J.~Zhang, Z.~Yin, S.~Zhang, X.~Huang, H.~Wang, T.~Wang,
  J.~Pang, W.~Ouyang, L.~Bai, W.~Zuo, L.-Y. Duan, D.~Zhou, and S.~Tang,
  ``{LabUtopia}: High-fidelity simulation and hierarchical benchmark for
  scientific embodied agents,'' in {\em Advances in Neural Information
  Processing Systems}, vol.~38, 2025.

\bibitem{wu2026labimus}
Y.~Wu, Z.~Jin, T.~Li, Y.~Zhang, Z.~Wang, S.~Wang, J.~Jiang, X.~Li, Y.~Zhang,
  J.~Tang, Z.~Che, and Y.~Xia, ``{Labimus}: A simulation and benchmark for
  humanoid dexterous manipulation in chemical laboratory,'' {\em arXiv preprint
  arXiv:2606.31037}, 2026.

\bibitem{lan2026autobio}
Z.~Lan, Y.~Jiang, R.~Wang, X.~Xie, R.~Zhang, Y.~Zhu, P.~Li, T.~Yang, T.~Chen,
  H.~Gao, X.~Yang, X.~Li, H.~Zhang, Y.~Mu, and P.~Luo, ``{AutoBio}: A
  simulation and benchmark for robotic automation in digital biology
  laboratory,'' in {\em International Conference on Learning Representations},
  2026.

\bibitem{abolhasani2023selfdriving}
M.~Abolhasani and E.~Kumacheva, ``The rise of self-driving labs in chemical and
  materials sciences,'' {\em Nature Synthesis}, vol.~2, pp.~483--492, 2023.

\bibitem{king2009automation}
R.~D. King, J.~Rowland, S.~G. Oliver, M.~Young, W.~Aubrey, E.~Byrne,
  M.~Liakata, M.~Markham, P.~Pir, L.~N. Soldatova, A.~Sparkes, K.~E. Whelan,
  and A.~Clare, ``The automation of science,'' {\em Science}, vol.~324,
  no.~5923, pp.~85--89, 2009.

\bibitem{zhao2023learning}
T.~Z. Zhao, V.~Kumar, S.~Levine, and C.~Finn, ``Learning fine-grained bimanual
  manipulation with low-cost hardware,'' in {\em Robotics: Science and
  Systems}, 2023.

\bibitem{chi2023diffusion}
C.~Chi, Z.~Xu, S.~Feng, E.~Cousineau, Y.~Du, B.~Burchfiel, R.~Tedrake, and
  S.~Song, ``Diffusion policy: Visuomotor policy learning via action
  diffusion,'' {\em The International Journal of Robotics Research}, vol.~44,
  no.~1, 2025.

\bibitem{black2025pi05}
K.~Black, N.~Brown, J.~Darpinian, K.~Dhabalia, D.~Driess, A.~Esmail, M.~Equi,
  C.~Finn, N.~Fusai, M.~Y. Galliker, D.~Ghosh, L.~Groom, K.~Hausman, B.~Ichter,
  S.~Jakubczak, T.~Jones, L.~Ke, D.~LeBlanc, S.~Levine, A.~Li-Bell,
  M.~Mothukuri, K.~Pertsch, A.~Z. Ren, L.~X. Shi, L.~Smith, J.~T. Springenberg,
  K.~Stachowicz, Q.~Vuong, H.~Walke, H.~Wang, L.~Yu, and U.~Zhilinsky,
  ``$\pi_{0.5}$: A vision-language-action model with open-world
  generalization,'' {\em arXiv preprint arXiv:2504.16054}, 2025.

\bibitem{chen2026robotwin2}
T.~Chen, Z.~Chen, B.~Chen, Z.~Cai, Y.~Liu, Z.~Li, Q.~Liang, X.~Lin, Y.~Ge,
  Z.~Gu, W.~Deng, Y.~Guo, T.~Nian, X.~Xie, Q.~Chen, K.~Su, T.~Xu, G.~Liu,
  M.~Hu, H.-a. Gao, K.~Wang, Z.~Liang, Y.~Qin, X.~Yang, P.~Luo, and Y.~Mu,
  ``Robotwin 2.0: A scalable data generator and benchmark with strong domain
  randomization for robust bimanual robotic manipulation,'' in {\em Proceedings
  of the 43rd International Conference on Machine Learning (ICML 2026)}, 2026.

\bibitem{liu2023libero}
B.~Liu, Y.~Zhu, C.~Gao, Y.~Feng, Q.~Liu, Y.~Zhu, and P.~Stone, ``{LIBERO}:
  Benchmarking knowledge transfer for lifelong robot learning,'' in {\em
  Advances in Neural Information Processing Systems}, vol.~36,
  pp.~44776--44791, 2023.

\bibitem{zhang2026r3dp}
Y.~Zhang, W.~Dong, Y.~Shi, Y.~Liang, J.~Gao, Q.~Yang, Y.~Lyu, Z.~Liang, Y.~Liu,
  C.~Xu, X.~Guo, W.~Sui, Y.~Jin, X.~Yang, Y.~Xu, and Y.~Mu, ``R3dp: Real-time
  3d-aware policy for embodied manipulation,'' in {\em European Conference on
  Computer Vision (ECCV)}, 2026.

\bibitem{bi2026motus}
H.~Bi, H.~Tan, S.~Xie, Z.~Wang, S.~Huang, H.~Liu, R.~Zhao, Y.~Feng, C.~Xiang,
  Y.~Rong, H.~Zhao, H.~Liu, Z.~Su, L.~Ma, H.~Su, and J.~Zhu, ``Motus: A unified
  latent action world model,'' in {\em Proceedings of the IEEE/CVF Conference
  on Computer Vision and Pattern Recognition}, 2026.

\bibitem{yuan2026fastwam}
T.~Yuan, Z.~Dong, Y.~Liu, and H.~Zhao, ``Fast-{WAM}: Do world action models
  need test-time future imagination?,'' {\em arXiv preprint arXiv:2603.16666},
  2026.

\bibitem{james2020rlbench}
S.~James, Z.~Ma, D.~Rovick~Arrojo, and A.~J. Davison, ``{RLBench}: The robot
  learning benchmark and learning environment,'' {\em IEEE Robotics and
  Automation Letters}, vol.~5, no.~2, pp.~3019--3026, 2020.

\bibitem{li2024behavior1k}
C.~Li, R.~Zhang, J.~Wong, C.~Gokmen, S.~Srivastava,
  R.~Mart{\'{i}}n-Mart{\'{i}}n, C.~Wang, G.~Levine, W.~Ai, B.~Martinez, H.~Yin,
  M.~Lingelbach, M.~Hwang, A.~Hiranaka, S.~Garlanka, A.~Aydin, S.~Lee, J.~Sun,
  M.~Anvari, M.~Sharma, D.~Bansal, S.~Hunter, K.-Y. Kim, A.~Lou, C.~R.
  Matthews, I.~Villa-Renteria, J.~H. Tang, C.~Tang, F.~Xia, Y.~Li, S.~Savarese,
  H.~Gweon, C.~K. Liu, J.~Wu, and L.~Fei-Fei, ``{BEHAVIOR-1K}: A
  human-centered, embodied {AI} benchmark with 1,000 everyday activities and
  realistic simulation,'' {\em arXiv preprint arXiv:2403.09227}, 2024.

\bibitem{jia2025discoverse}
Y.~Jia, G.~Wang, Y.~Dong, J.~Wu, Y.~Zeng, H.~Lin, Z.~Wang, H.~Ge, W.~Gu,
  K.~Ding, Z.~Yan, Y.~Cheng, Y.~Li, Z.~Wang, C.~Li, W.~Sui, L.~Shi, G.~Tian,
  R.~Huang, and G.~Zhou, ``{DISCOVERSE}: Efficient robot simulation in complex
  high-fidelity environments,'' in {\em 2025 IEEE/RSJ International Conference
  on Intelligent Robots and Systems}, 2025.

\bibitem{zakka2026mjlab}
K.~Zakka, Q.~Liao, B.~Yi, L.~Le~Lay, K.~Sreenath, and P.~Abbeel, ``{mjlab}: A
  lightweight framework for {GPU}-accelerated robot learning,'' {\em arXiv
  preprint arXiv:2601.22074}, 2026.

\bibitem{reeves1983particle}
W.~T. Reeves, ``Particle systems---a technique for modeling a class of fuzzy
  objects,'' {\em ACM Transactions on Graphics}, vol.~2, no.~2, pp.~91--108,
  1983.

\bibitem{wei2024dro}
Z.~Wei, Z.~Xu, J.~Guo, Y.~Hou, C.~Gao, Z.~Cai, J.~Luo, and L.~Shao,
  ``$\mathcal{D}(\mathcal{R}, \mathcal{O})$ grasp: A unified representation of
  robot and object interaction for cross-embodiment dexterous grasping,'' in
  {\em 2025 IEEE International Conference on Robotics and Automation (ICRA)},
  pp.~4982--4988, 2025.

\bibitem{fei2025trograspefficientgraph}
X.~Fei, Z.~Xu, H.~Fang, T.~Zhang, and L.~Shao, ``$\mathcal{T}(\mathcal{R},
  \mathcal{O})$ grasp: Efficient graph diffusion of robot-object spatial
  transformation for cross-embodiment dexterous grasping,'' {\em arXiv preprint
  arXiv:2510.12724}, 2025.

\bibitem{chen2025dexonomy}
J.~Chen, Y.~Ke, L.~Peng, and H.~Wang, ``{Dexonomy}: Synthesizing all dexterous
  grasp types in a grasp taxonomy,'' {\em Robotics: Science and Systems}, 2025.

\bibitem{yin2025lightninggrasp}
Z.-H. Yin and P.~Abbeel, ``Lightning grasp: High performance procedural grasp
  synthesis with contact fields,'' {\em arXiv preprint arXiv:2511.07418}, 2025.

\bibitem{hintjens2013zeromq}
P.~Hintjens, {\em {ZeroMQ}: Messaging for Many Applications}.
\newblock O'Reilly Media, 2013.

\end{thebibliography}

\clearpage
\newlength{\taskrowheight}
\setlength{\taskrowheight}{48pt}

\newcommand{\tasknamecell}[1]{%
    \begin{minipage}[c][\taskrowheight][c]{\linewidth}\centering
    #1
    \end{minipage}%
}

\newcommand{\taskbackgroundcell}[1]{%
    \begin{minipage}[c][\taskrowheight][c]{\linewidth}\centering
    #1
    \end{minipage}%
}

\newcommand{\taskframes}[1]{%
    \begin{minipage}[c][\taskrowheight][c]{\linewidth}\centering
    \includegraphics[height=45pt]{images/#1_1.png}\hspace{1pt}%
    \includegraphics[height=45pt]{images/#1_2.png}\hspace{1pt}%
    \includegraphics[height=45pt]{images/#1_3.png}\hspace{1pt}%
    \includegraphics[height=45pt]{images/#1_4.png}
    \end{minipage}%
}

\newcommand{\tasklanguagecell}[1]{%
    \begin{minipage}[c][\taskrowheight][c]{\linewidth}\raggedright
    #1
    \end{minipage}%
}

\appendix

\subsection{Dataset Structure}
Each demonstration in our dataset is stored as a time-aligned trajectory
$D = \{(\mathbf{o}_t, \mathbf{a}_t)\}_{t=1}^{T}$. At each time step, the observation contains
multi-view camera images, robot proprioception, and a language instruction:
\begin{equation}
    \begin{aligned}
    \mathbf{o}_t &= \{o_{\mathrm{cam},t}, o_{\mathrm{prop},t}, o_{l,t}\}, \\
    o_{\mathrm{cam},t} &= \{I_t^{\mathrm{front}}, I_t^{\mathrm{wrist1}},
    I_t^{\mathrm{left/wrist2}}\}, \\
    o_{\mathrm{prop},t} &= \{q_{\mathrm{arm},t}, q_{\mathrm{hand/gripper},t}\}.
    \end{aligned}
\end{equation}
Here, the superscript identifies the camera view. For the Tube Transfer (XHand) task, only the
front and wrist1 views, $(I_t^{\mathrm{front}}, I_t^{\mathrm{wrist1}})$, are included to match the
real-robot setup. The proprioceptive vector $q_{\mathrm{arm},t}$ contains the manipulator joint
positions, while $q_{\mathrm{hand/gripper},t}$ contains the joint positions of the end-effector
hand or gripper. The latter is an articulated hand in dexterous setups and a gripper in simpler
setups. The term $o_{l,t}$ denotes the language instruction associated with the trajectory.
The corresponding action at time step $t$ is stored as
\begin{equation}
    \mathbf{a}_t
    = \left[ pos_{\mathrm{ee},t}, quat_{\mathrm{ee},t},
    q_{\mathrm{hand/gripper},t} \right],
\end{equation}
where $pos_{\mathrm{ee},t} \in \mathbb{R}^{3}$ is the end-effector position,
$quat_{\mathrm{ee},t} \in \mathbb{R}^{4}$ is a unit quaternion representing the end-effector
orientation, and $q_{\mathrm{hand/gripper},t} \in
\mathbb{R}^{n_{\mathrm{hand/gripper}}}$ is the hand- or gripper-joint position vector. For
dual-arm tasks, the action contains the concatenated commands for both arms:
\begin{equation}
    \mathbf{a}_{\mathrm{dual},t} = \mathbf{a}_{1,t} \oplus \mathbf{a}_{2,t}.
\end{equation}

\subsection{Visualization and Language Instruction of DexBioLab Tasks}

Table~\ref{tab:appendix-task-instructions} presents the visualization information
and language instructions for the fourteen DexBioLab tasks. 

\begin{table*}[t]
\centering
\caption{Visualization and language instruction of the fourteen DexBioLab tasks.}
\label{tab:appendix-task-instructions}
\small
\fontsize{9}{9.5}\selectfont
\renewcommand{\arraystretch}{1.16}
\setlength{\tabcolsep}{3pt}
\begin{tabular*}{\textwidth}{@{\extracolsep{\fill}}>{\centering\arraybackslash}m{0.11\textwidth}>{\centering\arraybackslash}m{0.11\textwidth}>{\centering\arraybackslash}m{0.49\textwidth}>{\raggedright\arraybackslash}m{0.24\textwidth}@{}}
\toprule
\textbf{Task} & \textbf{Gaussian Background} & \textbf{Visualization} & \textbf{Language Instruction} \\
\midrule
\tasknamecell{Attach Pipette Tip} & \taskbackgroundcell{Fume Hood} & \taskframes{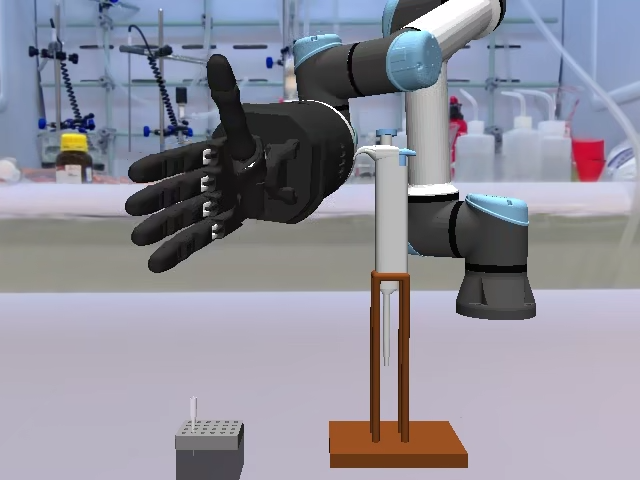} & \tasklanguagecell{Attach the pipette tip with the Shadow hand.} \\\midrule
\tasknamecell{Aspirate with Pipette} & \taskbackgroundcell{Fume Hood} & \taskframes{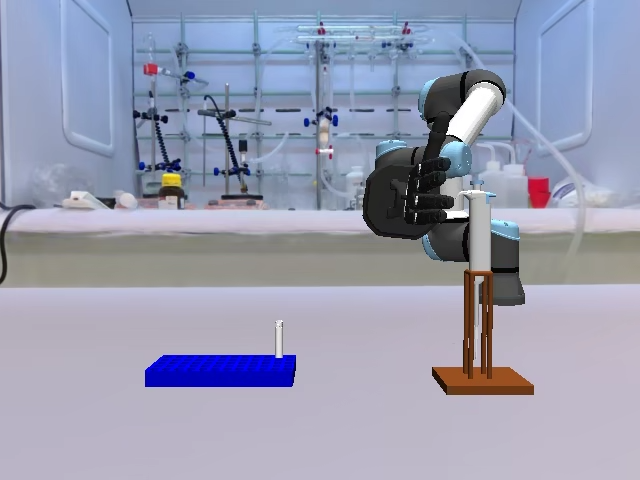} & \tasklanguagecell{Pick up the pipette and move it directly above the cryovial to perform the pipetting operation.} \\\midrule
\tasknamecell{Transfer Tube (XHand)} & \taskbackgroundcell{Real Tube-Rack Scene} & \taskframes{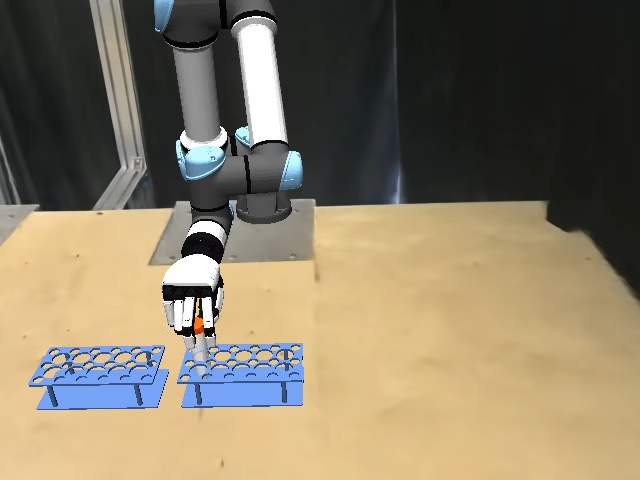} & \tasklanguagecell{Pick up the centrifuge tube and move it to the other rack, row [row], col [column].} \\\midrule
\tasknamecell{Manipulate Tweezers} & \taskbackgroundcell{Lab Bench} & \taskframes{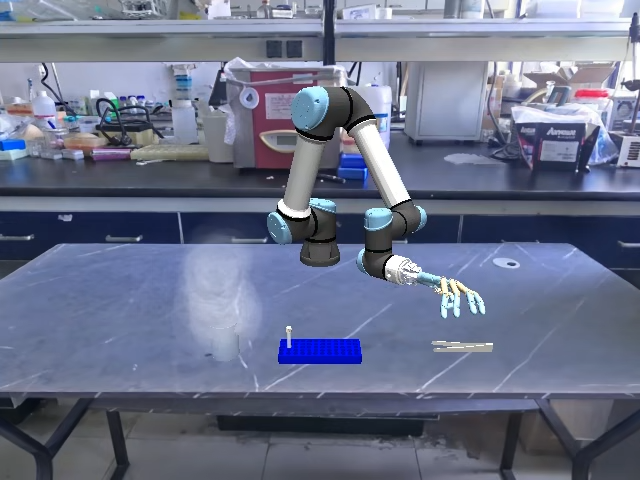} & \tasklanguagecell{Pick up the tweezers, move them to the cryovial, grasp it with the tweezers, and place it into the fogging liquid nitrogen flask.} \\\midrule
\tasknamecell{Heat \& Stir Agar Solution} & \taskbackgroundcell{Lab Bench} & \taskframes{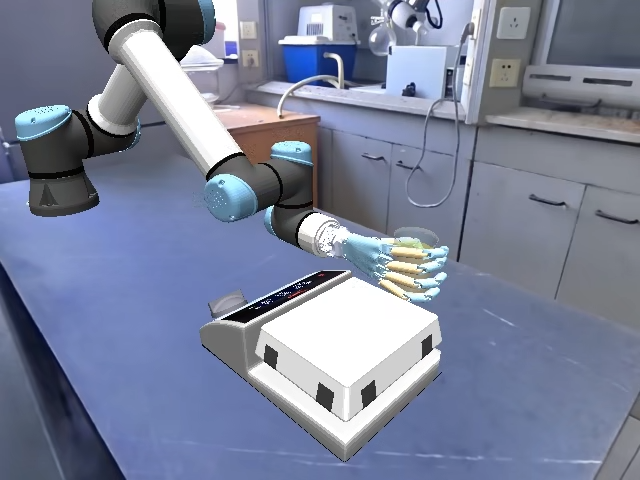} & \tasklanguagecell{Place the beaker on the heating plate, turn the knob on, then return the knob to its home position with the Wuji hand.} \\\midrule
\tasknamecell{Close Centrifuge Lid} & \taskbackgroundcell{Centrifuge Station} & \taskframes{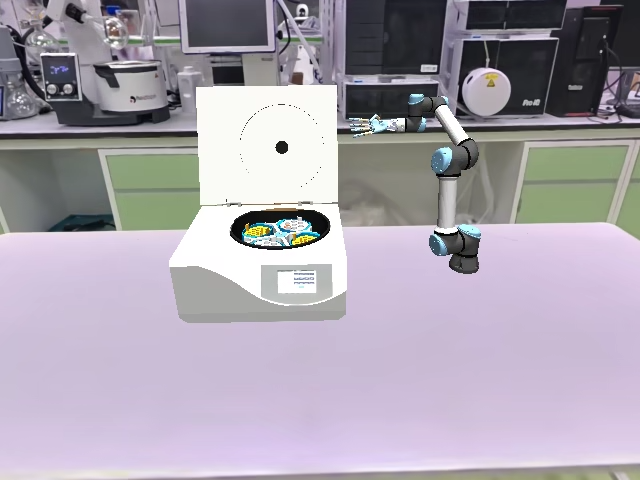} & \tasklanguagecell{Close the centrifuge lid with the Wuji hand.} \\\midrule
\tasknamecell{Load Centrifuge} & \taskbackgroundcell{Centrifuge Station} & \taskframes{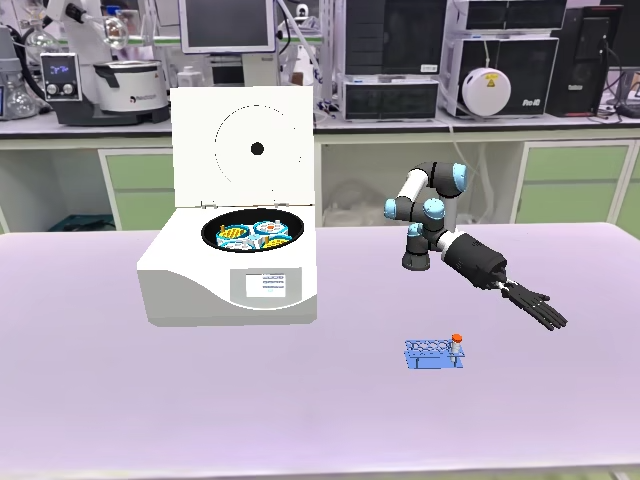} & \tasklanguagecell{Insert the tube into the centrifuge slot symmetrically opposite the already placed tube with the Shadow hand.} \\\midrule
\tasknamecell{Tighten Tube Cap} & \taskbackgroundcell{Autolab Enclosure} & \taskframes{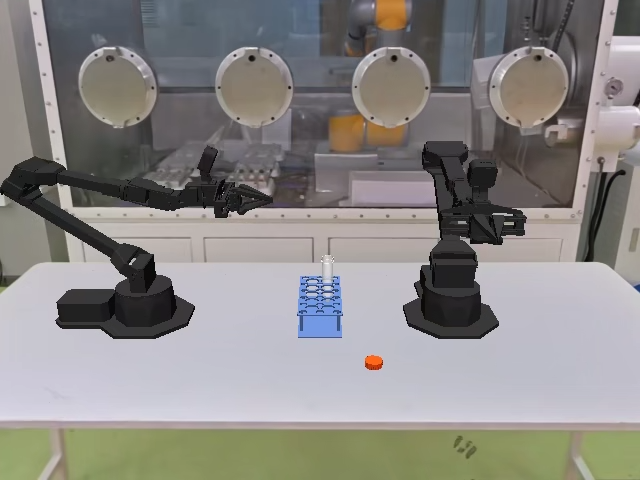} & \tasklanguagecell{Dual-Aloha arms screwing on centrifuge tube cap: one grasps the tube from the rack and then grips the tube while the other twists the cap.} \\\midrule
\tasknamecell{Transfer Tube (Shadow)} & \taskbackgroundcell{Autolab Enclosure} & \taskframes{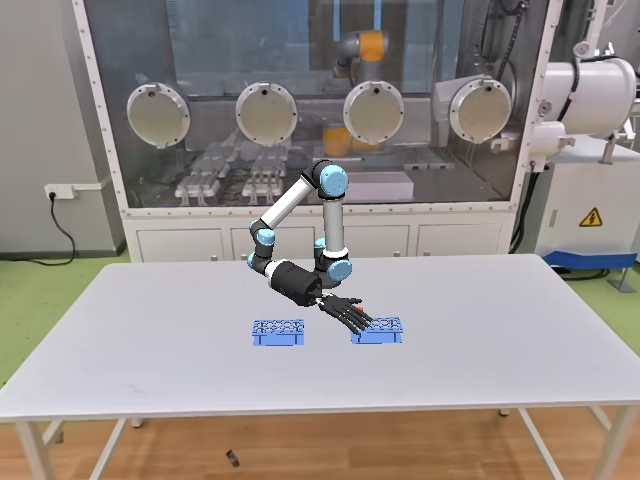} & \tasklanguagecell{Insert the tube into rack row [row], column [column] with the Shadow hand.} \\\midrule
\tasknamecell{Ignite Bunsen Burner} & \taskbackgroundcell{Heating and Mixing Station} & \taskframes{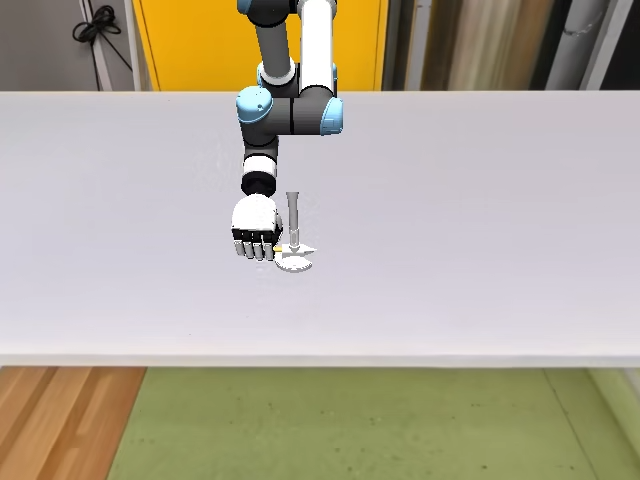} & \tasklanguagecell{Rotate the Bunsen burner's silver ring to turn it on, then rotate the yellow ring to increase the flame temperature.} \\
\midrule
\tasknamecell{Loosen Tube Cap} & \taskbackgroundcell{Autolab Enclosure} & \taskframes{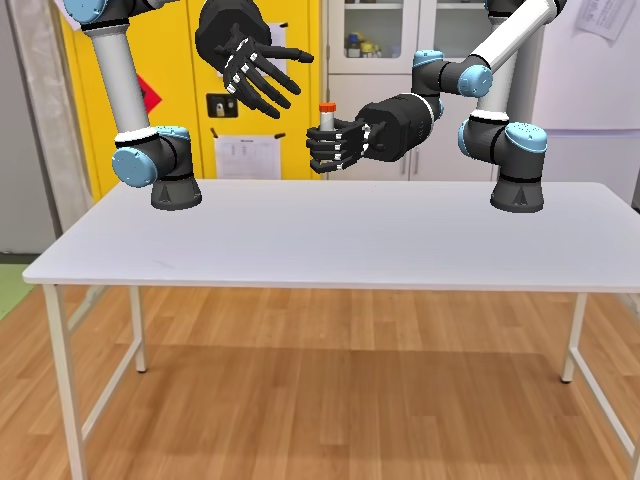} & \tasklanguagecell{dual-Shadow unscrewing centrifuge tube cap: one stabilizes tube while the other twists cap} \\
\bottomrule
\multicolumn{4}{r}{\textit{Continued on next page}}\\[-2pt]
\end{tabular*}
\end{table*}

\begin{table*}[t]
\centering
\addtocounter{table}{-1}
\caption[]{Visualization and language instruction of the fourteen DexBioLab tasks (continued).}
\small
\fontsize{9}{9.5}\selectfont
\renewcommand{\arraystretch}{1.16}
\setlength{\tabcolsep}{3pt}
\begin{tabular*}{\textwidth}{@{\extracolsep{\fill}}>{\centering\arraybackslash}m{0.11\textwidth}>{\centering\arraybackslash}m{0.11\textwidth}>{\centering\arraybackslash}m{0.49\textwidth}>{\raggedright\arraybackslash}m{0.24\textwidth}@{}}
\toprule
\textbf{Task} & \textbf{Gaussian Background} & \textbf{Visualization} & \textbf{Language Instruction} \\
\midrule
\tasknamecell{Open Thermal Cycler Lid} & \taskbackgroundcell{Reagent Cabinet} & \taskframes{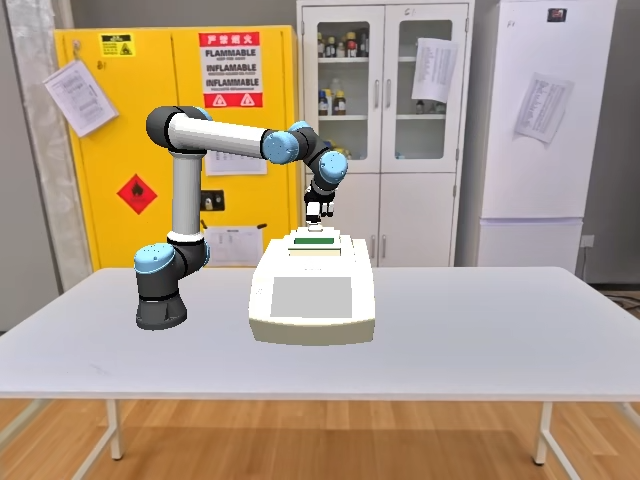} & \tasklanguagecell{Open the thermal cycler with the XHand.} \\\midrule
\tasknamecell{Close Thermal Cycler Lid} & \taskbackgroundcell{Reagent Cabinet} & \taskframes{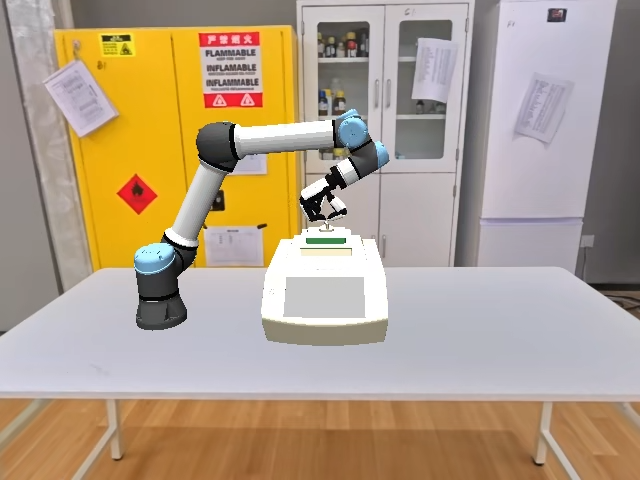} & \tasklanguagecell{Close the thermal cycler with the XHand.} \\\midrule
\tasknamecell{Operate Mixer Panel} & \taskbackgroundcell{Heating and Mixing Station} & \taskframes{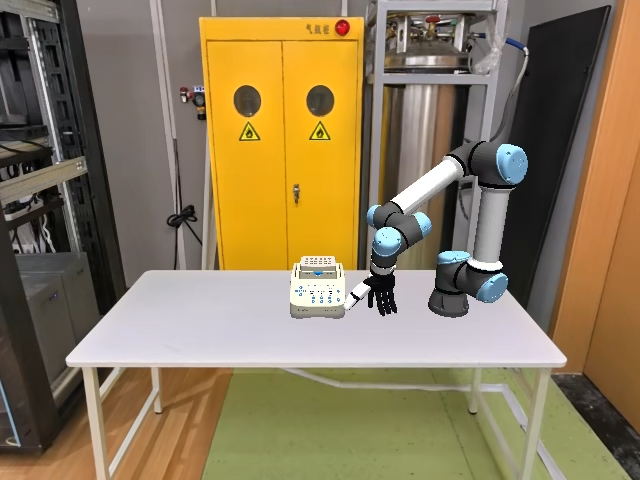} & \tasklanguagecell{Set the mixer to [time] seconds, [temperature] degrees Celsius, and [speed] RPM with the XHand.} \\
\bottomrule
\end{tabular*}
\end{table*}

\subsection{Randomization Settings of DexBioLab Tasks}

At the beginning of each episode, DexBioLab randomizes both the task scene and
the robot initialization. Scene randomization covers object poses, discrete
slot assignments, and, where applicable, visual appearance. Robot
randomization perturbs the initial joint configuration within task-specific
ranges. Table~\ref{tab:appendix-randomization} reports the sampling rules and
reset ranges used for each task.
\begin{table*}[t]
\centering
\caption{Task-specific randomization settings for the fourteen DexBioLab tasks.}
\label{tab:appendix-randomization}
\small
\renewcommand{\arraystretch}{1.22}
\setlength{\emergencystretch}{3em}
\setlength{\tabcolsep}{4pt}
\setlength{\extrarowheight}{2pt}
\begin{tabular*}{\textwidth}{@{\extracolsep{\fill}}>{\raggedright\arraybackslash}m{0.22\textwidth}>{\raggedright\arraybackslash}m{0.42\textwidth}>{\raggedright\arraybackslash}m{0.28\textwidth}@{}}
\toprule
\textbf{Task} & \textbf{Object / visual randomization} & \textbf{Robot reset ranges (rad.)} \\
\midrule
Attach Pipette Tip & Tip slot sampled from the $6\times4$ tip-box grid. & UR5e: $[-0.20,0.20]$. \\
\midrule
\begin{minipage}[c][2.6\baselineskip][c]{\linewidth}\raggedright Aspirate with Pipette\end{minipage} & \begin{minipage}[c][2.6\baselineskip][c]{\linewidth}\raggedright Cryovial slot sampled from the $5\times12$ rack grid.\end{minipage} & \begin{minipage}[c][2.6\baselineskip][c]{\linewidth}\raggedright UR5e: $[-1.89,2.76]$;\\[-1pt] Shadow: $[-0.95,1.12]$.\end{minipage} \\
\midrule
\begin{minipage}[c][4.4\baselineskip][c]{\linewidth}\raggedright Transfer Tube\end{minipage} &
\begin{minipage}[c][4.4\baselineskip][c]{\linewidth}\raggedright XHand: source tube slot sampled from the $5\times2$ rack.\\[2pt] Shadow: target row $\in\{0,\ldots,4\}$ and column $\in\{0,1\}$ sampled; tube pose follows the selected slot.\end{minipage} &
\begin{minipage}[c][4.4\baselineskip][c]{\linewidth}\raggedright UR5e (XHand): $[-0.20,0.30]$;\\[1pt] UR5e (Shadow): $[-1.89,2.07]$;\\[1pt] Shadow: $[-0.95,1.12]$.\end{minipage} \\
\midrule
 \begin{minipage}[c][2.6\baselineskip][c]{\linewidth}\raggedright Manipulate Tweezers\end{minipage} & \begin{minipage}[c][2.6\baselineskip][c]{\linewidth}\raggedright Tweezers $dx,dy\sim U(-0.025,0.025)$ m and yaw $\sim U(-0.10,0.10)$ rad; tube slot sampled.\end{minipage} & \begin{minipage}[c][2.6\baselineskip][c]{\linewidth}\raggedright Wuji: $[-0.12,0.12]$.\end{minipage} \\
\midrule
Heat \& Stir Agar Solution & \begin{minipage}[c]{\linewidth}\raggedright\mbox{Plate: $dx\sim U(-0.02,0.02)$, $dy\sim U(0,0.02)\,\mathrm{m}$};\\[-1pt] \mbox{beaker: $dx\sim U(-0.02,0.02)$, $dy\sim U(-0.02,0)\,\mathrm{m}$.}\end{minipage} & UR5e: $[-0.10,0.10]$. \\
\midrule
Close Centrifuge Lid & Centrifuge $dx,dy\sim U(-0.025,0.025)$ m. & UR5e: $[-0.10,0.10]$. \\
\midrule
\begin{minipage}[c][2.6\baselineskip][c]{\linewidth}\raggedright Load Centrifuge\end{minipage} & \begin{minipage}[c][2.6\baselineskip][c]{\linewidth}\raggedright Target slot sampled from the $5\times2$ safe-slot set; tube pose follows the selected slot.\end{minipage} & \begin{minipage}[c][2.6\baselineskip][c]{\linewidth}\raggedright UR5e: $[-1.19,1.07]$;\\[-1pt] Shadow: $[-0.95,1.12]$.\end{minipage} \\
\midrule
Tighten Tube Cap & Cap $dx,dy\sim U(-0.01,0.01)$ m; rack slot sampled. & Aloha: $[-0.05,0.08]$. \\
\midrule
 \begin{minipage}[c][2.6\baselineskip][c]{\linewidth}\raggedright Ignite Bunsen Burner\end{minipage} & \begin{minipage}[c][2.6\baselineskip][c]{\linewidth}\raggedright Burner $dx\sim U(-0.07,0.07)$, $dy\sim U(-0.055,0.055)$ m; flame appearance randomized.\end{minipage} & \begin{minipage}[c][2.6\baselineskip][c]{\linewidth}\raggedright UR5e: $[-0.08,0.08]$;\\[-1pt] XHand: $[-0.05,0.05]$.\end{minipage} \\
\midrule
Loosen Tube Cap & Cap $dx,dy\sim U(-0.01,0.01)$ m; rack slot sampled. & Shadow: $[-0.95,1.12]$. \\
\midrule
Open Thermal Cycler Lid & -- & -- \\
\midrule
Close Thermal Cycler Lid & Cycler $dx\sim U(-0.08,0)$ m. & UR5e: $[-0.16,0.16]$. \\
\midrule
Operate Mixer Panel & -- & UR5e: $[-0.20,0.30]$. \\
\bottomrule
\end{tabular*}
\par\smallskip
\end{table*}

\subsection{Evaluation Infrastructure}
DexBioLab provides a unified evaluation infrastructure built around domain randomization,
task-specific success checking, and model-independent policy communication. At each episode reset,
DexBioLab applies domain randomization to the robot's initial joint configuration and the object poses.
Each task is equipped with a task-specific success-checking function that determines completion from
the simulator state. DexBioLab and the policy runtime are decoupled via a
ZeroMQ~\cite{hintjens2013zeromq} observation--action interface, allowing policies to run in
independent Python environments with different versions and dependencies. The simulator serializes
the current visual observations, proprioceptive state, and language prompt to an external inference
process and receives action chunks in return. Consequently, a model can be evaluated from a separate
process, machine, or software environment without being integrated into the simulator's dependency
stack.

The thread mechanism in DexBioLab is inherited from AutoBio~\cite{lan2026autobio}.
We ported its original MuJoCo implementation to mjlab, enabling large-scale parallel
simulation from CPU to GPU while preserving the mechanism's behavior.

\subsection{Training Details}
All evaluated models are fine-tuned. For the simulated-task evaluation, each
model is trained for 30,000 optimization steps on a single NVIDIA H200 GPU. The
batch size for $\pi_{0.5}$ is 32.

For the real-world Transfer Tube (XHand) experiment, we train the
$\pi_{0.5}$ model on 100 real-world demonstrations. The training uses the same
30-step action horizon as the simulated $\pi_{0.5}$ experiments. We train for
30,000 optimization steps on a single GPU with a batch size of 32. The peak
learning rate is
$5\times10^{-5}$ with a 3,333-step warmup; the optimizer uses AdamW with
$\beta_1=0.9$, $\beta_2=0.95$, weight decay $10^{-10}$, and gradient-norm
clipping at 1.0.

\begin{figure*}[t]
    \centering
    \includegraphics[height=8.8in,keepaspectratio]{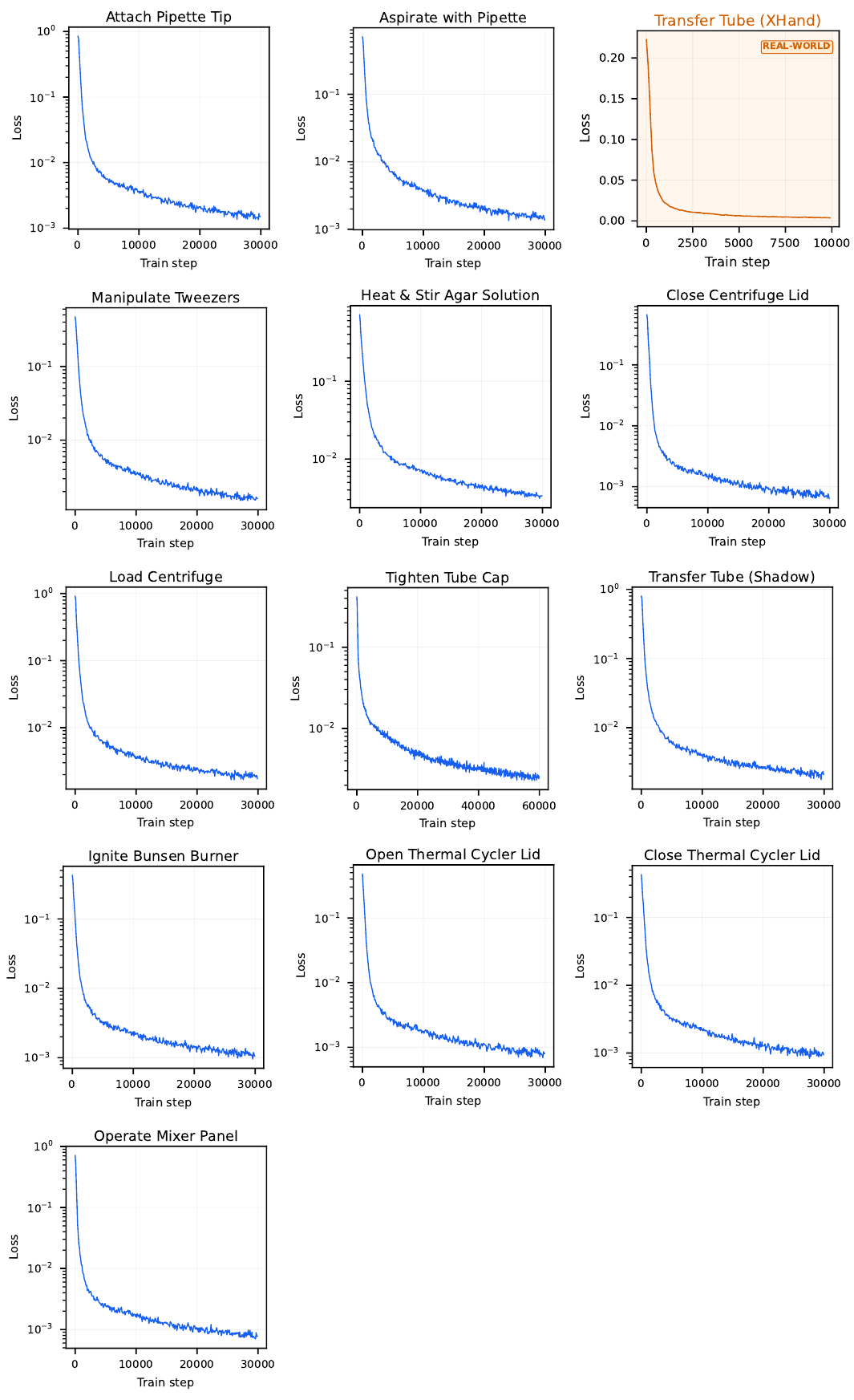}
    \caption{Raw training loss curves for 13 tasks. The orange panel denotes
    $\pi_{0.5}$ training on real-world data for Transfer Tube (XHand),
    whereas the blue panels denote $\pi_{0.5}$ training on simulated task data.}
    \label{fig:pi05-training-loss}
\end{figure*}

Figure~\ref{fig:pi05-training-loss} shows the raw training losses for the 12
simulated tasks and the real-world Transfer Tube (XHand) experiment.

\end{document}